\documentclass[10pt, letterpaper, journal, twocolumn]{IEEEtran}

\usepackage[utf8]{inputenc}
\usepackage{graphicx,amsmath,amsfonts,amssymb,amsthm,stmaryrd,times}
\usepackage{graphicx,psfrag}
\usepackage[T1]{fontenc}
\usepackage[all]{xy}
\usepackage{multirow}
\usepackage{rotating}
\usepackage{xcolor}
\usepackage{wasysym}
\usepackage{url}
\usepackage{type1cm}
\usepackage{lettrine}
\usepackage{epstopdf}
\theoremstyle{definition}

\newtheorem{exmp}{Example}
\newtheorem*{rmk}{Remark}

\def\amax{\kern 0em\hbox{\rm \kern .25em\lower.1ex\hbox{\rlap{$\vee$}}\kern -.07em\lower.2ex\hbox{$\square$}\kern.25em}}
\def\amin{\kern 0em\hbox{\rm \kern .25em\lower.1ex\hbox{\rlap{$\wedge$}}\kern -.07em\lower.2ex\hbox{$\square$}\kern.25em}}

\def\boxmax{\kern 0em\hbox{\rm \kern .25em\lower.1ex\hbox{\rlap{$\vee$}}\kern -.07em\lower.2ex\hbox{$\square$}\kern.25em}}
\def\boxmin{\kern 0em\hbox{\rm \kern .25em\lower.1ex\hbox{\rlap{$\wedge$}}\kern -.07em\lower.2ex\hbox{$\square$}\kern.25em}}
\def\dualimp{\kern 0em\hbox{\rm \kern .25em\lower.1ex\hbox{\rlap{$\Rightarrow$}}\kern 0em\lower-1.2ex\hbox{$\overline{\hspace{2ex}}$}\kern.25em}}

\def\circmax{\kern 0em\hbox{\rm \kern .25em\lower.1ex\hbox{\rlap{$\vee$}}\kern -.18em\lower-.1ex\hbox{$\bigcirc$}\kern.25em}}
\def\circmin{\kern 0em\hbox{\rm \kern .25em\lower.1ex\hbox{\rlap{$\wedge$}}\kern -.18em\lower-.0ex\hbox{$\bigcirc$}\kern.25em}}

\newcommand{\vetx}{\mathbf{x}}

\newcommand{\ii}{\mathbf{i}}
\newcommand{\jj}{\mathbf{j}}
\newcommand{\kk}{\mathbf{k}}
\newcommand{\quat}[1]{{#1}_0 + {#1}_1 \ii + {#1}_2 \jj + {#1}_3 \kk}
\newcommand{\re}[1]{\text{Re}\left\{#1\right\}}
\newcommand{\ve}[1]{\text{Ve}\left\{#1\right\}}

\newcommand{\bb}{\begin{equation}}
\newcommand{\ee}{\end{equation}}

\title{On the Dynamics of Hopfield Neural Networks on Unit Quaternions}
\author{Marcos Eduardo Valle\thanks{Marcos Eduardo Valle is with the Department of Applied Mathematics, University of Campinas, Brazil, Campinas, SP, Brazil. (email:  valle@ime.unicamp.br).} and Fidelis Zanetti de Castro\thanks{Fidelis Zanetti de Castro is with the Federal Institute of Education, Science and Technology of Esp\'irito Santo at Serra, Serra, ES, Brazil. (email: fidelis@ifes.edu.br)}
\thanks{This work was supported in part by CNPq under grant no. 305486/2014-4.}
}

\begin{document}

\newpage
\thispagestyle{empty}
\begin{minipage}{0.8\textwidth}
\noindent {\Huge  IEEE Copyright Notice} \vspace{2cm}

\noindent Personal use of this material is permitted.  Permission from IEEE must be obtained for all other uses, in any current or future media, including reprinting/republishing this material for advertising or promotional purposes, creating new collective works, for resale or redistribution to servers or lists, or reuse of any copyrighted component of this work in other works.
\vspace{1cm}

\noindent This file corresponds to the accepted version of the manuscript published in IEEE TRANSACTIONS ON NEURAL NETWORKS AND LEARNING SYSTEMS, VOL. 29, NO. 6, JUNE 2018. Digital Object Identifier 10.1109/TNNLS.2017.2691462 
\end{minipage}

\maketitle

\begin{abstract}
In this paper, we first address the dynamics of the elegant multi-valued quaternionic Hopfield neural network (MV-QHNN) proposed by Minemoto and collaborators. 
Contrary to what was expected, we show that the MV-QHNN, as well as one of its variation, does not always come to rest at an equilibrium state under the usual conditions. In fact, we provide simple examples in which the network yields a periodic sequence of quaternionic state vectors. Afterward, we turn our attention to the continuous-valued quaternionic Hopfield neural network (CV-QHNN), which can be derived from the MV-QHNN by means of a limit process. The CV-QHNN can be implemented more easily than the MV-QHNN model. Furthermore, the asynchronous CV-QHNN always settles down into an equilibrium state under the usual conditions. Theoretical issues are all illustrated by examples in this paper. 
\end{abstract}


\section{Introduction}
\lettrine{T}{he} Hopfield neural network (HNN) is one of the most important neural network with applications in different areas, including signal reconstruction \cite{perry00}, image analysis \cite{sun00}, and optimization \cite{hopfield85,serpen08}. Although the HNN has been initially conceived for binary state neurons \cite{hopfield82}, it have been extended to multistate neurons over the years using complex \cite{jankowski96,muezzinoglu03,lee06,garimella16}, quaternions \cite{isokawa08b,isokawa13,valle14bracis,kobayashi15,minemoto16,Kobayashi16a,Kobayashi17b}, and many other hyper-complex algebras \cite{vallejo08,kobayashi13,kuroe16,popa16c}.

In contrast to real-valued neural networks, hyper-complex models treat multi-dimensional data as a single entity \cite{nitta09,hirose12,kobayashi17a}. In particular, quaternion-valued neural networks are designed to process four dimensional data. Applications of quaternion-valued networks include control \cite{fortuna96,arena98}, signal and image processing \cite{minemoto16,isokawa03,jahanchahi14,xia15,xu16,hikosaka16}, classification and prediction \cite{ujang11,mandic11,shang14,talebi15,popa16a}. 
%
%

In this paper, we focus on quaternionic Hopfield neural networks (QHNNs), a topic that has been extensively investigated in the last years \cite{isokawa13,valle14bracis,minemoto16,isokawa12,osana12,minemoto15a,valle16wcci}. The reader interested on a comprehensive review on QHNNs is invited to consult \cite{isokawa13}.
Briefly, the main difference between the several QHNN models resides in the activation function and its target set -- which comprehend all possible states of a neuron. For instance, a broad class of QHNN models, referred to as split QHNNs, are derived by applying a real-valued function, such as the sign or $\tanh$ functions, to each component of the activation potential of a quaternion-valued neuron \cite{isokawa06,isokawa07,isokawa08a}. In this paper, we shall confine our attention on the particular class of Hopfield neural networks whose states are represented by unit quaternions. 


As far as we know, the first QHNN model on unit quaternions have been introduced by Isokawa et al. in 2008 \cite{isokawa08b,isokawa13}. This network uses the phase-angle representation of unit quaternions and an extension of the signum function. Recently, we pointed out in a conference paper that the quaternionic signum function proposed by Isokawa et al. is computationally unstable \cite{valle16wcci}. Furthermore, we provided an example in which the energy of the network increases after a change of states even if we work in exact arithmetic. As a consequence, contrary to what have been thought, we cannot ensure that the MV-QHNN of Isokawa always comes to rest at a stable stationary state. 

Recently, Minemoto et al. introduced a modified version of the quaternionic multivalued signum function, which is numerically stable \cite{minemoto16,valle16wcci}. In few words, the model of Minemoto et al. is obtained by shifting the phase-angles of the previous model of Isokawa and collaborators. In spite of the numerical stability, we remarked in the conference paper \cite{valle16wcci} that the usual energy function of the MV-QHNN of Minemoto does not necessarily decrease. In this paper, we address the subtle assumptions which have been wrongly accepted as true to assert the convergence of a sequence produced by the MV-QHNN of Minemoto et al. \cite{isokawa08b,isokawa13}. We believe that this is an important theoretical issue because MV-QHNN models have been used as the basis for other quaternionic associative memories \cite{minemoto16,minemoto15a}. 

In this paper, we also address a modified version of the MV-QHNN of Minemoto in which all phase-angles of a neuron are updated simultaneously. The modified version of the MV-QHNN of Minemoto is referred to as the MV-QHNN3. In our conference paper \cite{valle16wcci}, we remarked that the sequence produced by the MV-QHNN3 converges to a stationary state under the usual conditions on the synaptic weight matrix. In this paper, however, we show that this conjecture is wrong. Precisely, we provide an example in which the MV-QHNN3, using either asynchronous or parallel update mode, yields a periodic sequence of quaternionic vectors.

Apart from the MV-QHNN models, in this paper we also consider a QHNN model in which the state of a neuron is obtained by setting its activation potential to length one \cite{valle14bracis,Kobayashi16a}. This model, called {\em continuous-valued QHNN} (CV-QHNN), corresponds to the limit of the MV-QHNN3 model when the number of states tends to infinity. Different from the MV-QHNNs, the CV-QHNN is not based on the phase-angle representation of unit quaternions. Thus, it can be analyzed and implemented more easily than the former models. Moreover, the CV-QHNN operating in an asynchronous update manner always come to rest at a stable equilibrium under the usual (quaternionic) conditions on the synaptic weights. 

This paper is organized as follows: Next section presents the mathematical background on quaternions. In Section \ref{sec:MV-QHNN}, we review the MV-QHNN models and focus on convergence issues. The CV-QHNN model is described in Section \ref{sec:CV-QHNN}. 
The paper finishes with concluding remarks in Section \ref{sec:concluding}.

\section{Mathematical background on quaternions} \label{sec:quat}
Quaternions, introduced in the late 19th century by Hamilton, are four dimensional hyper-complex numbers, which generalize real and complex numbers \cite{arena98}.

A quaternion $q$ can be seen as a 4-tuple of real numbers, i.e., $q=(q_0,q_1,q_2,q_3)$. A quaternion $q$ can also be represented using the algebraic form
\bb q = q_0 + q_1 \ii + q_2 \jj + q_3 \kk, \ee
where $\ii, \jj$, and $\kk$, typed in this paper using boldface letters, denote hyper-imaginary units that satisfy the rules
\bb \ii^2 = \jj^2 = \kk^2 = \ii \jj \kk = -1. \ee
Furthermore, a quaternion $q$ can be represented as
\[ q = q_0+\vec{q}, \] 
where $q_0$ and $\vec{q}=q_1 \ii + q_2 \jj + q_3 \kk$ are called, respectively, the real part and the vector part of $q$. We denote the real and the vector part of $q$ by $\re{q}:=q_0$ and $\ve{q}:=\vec{q}$, respectively.

The addition of quaternions is defined analogously to the complex numbers, that is, by adding the respective components. Formally, the sum of $p = \quat{p}$ and $q=\quat{q}$ is given by 
\[ p+q=(p_0+q_0)+(p_1+q_1)\ii+(p_2+q_2)\jj+(p_3+q_3)\kk. \]
The product of $p$ and $q$, however, is the quaternion 
\[ pq=p_0 q_0 - \vec{p} \cdot \vec{q} +p_0 \vec{q} + q_0 \vec{p} + \vec{p} \times \vec{q},\]
where $\vec{p} \cdot \vec{q}$ and $\vec{p} \times \vec{q}$ denote, respectively, the usual scalar and cross products between $\vec{p}$ and $\vec{q}$. We call the reader's attention to the fact that the quaternion product is associative but it is not commutative. 

Like complex numbers, the conjugate $\bar{q}$ and the norm $|q|$ of a quaternion $q$ are given respectively by 
\[ \bar{q} = q_0-\vec{q} \quad \mbox{and} \quad |q| = \sqrt{\bar{q} q} =  \sqrt{q_0^2 + q_1^2 + q_2^2 +q_3^2}. \]
We say that $q$ is a {\em unit} quaternion when $|q|=1$. We denote by $\mathbb{S}$ the set of all unit quaternions, i.e., $\mathbb{S} = \{q \in \mathbb{H}: |q|=1\}$. 

Note that, for any quaternions $p=\quat{p}$ and $q=\quat{q}$, we have
\[ \re{\bar{q}p} = q_0p_0+q_1p_1+q_2p_2+q_3p_3. \]
Hence, from Cauchy-Schwarz inequality, the angle $A \in [0,\pi]$ between $p$ and $q$ satisfies
\bb \label{eq:cossenos} \re{\bar{q}p} = |q||p| \cos(A).\ee
We say that $p$ and $q$ are parallel if $|\cos(A)|=1$.

Alternatively, a quaternion can be expressed using the {\em phase-angle} representation, which is derived from the relationship between quaternions and rotations in $\mathbb{R}^3$ \cite{buelow99}. In the phase-angle representation, a quaternion $q$ is written as
\bb \label{eq:phase-angle} q = |q| e^{\ii \phi} e^{\kk \psi} e^{\jj \theta}, \ee
where $\phi \in \left[-\pi, \pi \right)$, $\theta \in \left[-\frac{\pi}{2},\frac{\pi}{2} \right)$, and $\psi \in \left[-\frac{\pi}{4},\frac{\pi}{4} \right]$. The exponential of an hyper-imaginary unit is defined using Euler's formula, i.e., $e^{\ii \phi} = \cos(\phi)+\ii \sin(\phi)$, $e^{\kk \psi} = \cos(\psi)+\kk \sin(\psi)$, and $e^{\jj \theta} = \cos(\theta)+\jj \sin(\theta)$. 

We would like to remark that not all quaternions have an unique phase-angle representation. In fact, the phase-angle representation of $q = 0$ is not unique. Furthermore, when $|\psi|=\frac{\pi}{4}$, the angles $\phi$ and $\theta$ are not uniquely determined \cite{buelow99}. This singularity is known in the theory of Euler matrices as \textit{Gimbal lock}. We define the set $\mathcal{A}$ of all quaternions that can be uniquely represented using \eqref{eq:phase-angle}. Formally, we have
\bb \label{eq:SetA} \mathcal{A} = \left \{q=|q| e^{\ii \phi} e^{\kk \psi} e^{\jj \theta}: q \neq 0 \; \mbox{and} \; |\psi| \neq \frac{\pi}{4} \right\}.\ee

\section{Multivalued Quaternionic Hopfield Neural Networks} \label{sec:MV-QHNN}

As far as we know, the first multivalued quaternionic Hopfield network (MV-QHNN) on unit quaternions have been proposed and analyzed by Isokawa et al. \cite{isokawa08b,isokawa13}. Briefly, using the phase-angle representation, Isokawa and collaborators defined a quaternionic multi-valued signum function which generalizes the complex-valued signum function of Jankowski et al. \cite{jankowski96}. It turns out that the quaternionic multi-valued signum function proposed initially by Isokawa and collaborators is numerically unstable \cite{valle16wcci}. Therefore, in this paper we only consider the multivalued QHNN (MV-QHNN) proposed recently by Minemoto et al. \cite{minemoto16,minemoto15a}. In few words, the MV-QHNN of Minemoto is obtained by a simple modification on the quaternionic multivalued signum function proposed initially by Isokawa and collaborators.

\subsection{MV-QHNN of Minemoto} \label{sec:Minemoto}

Like the traditional discrete-time Hopfield neural network, the MV-QHNN is a single-layer recursive neural network. Let $w_{ij}$ denotes the $j$th quaternionic synaptic weight of the $i$th neuron of a network with $n$ neurons. Also, let $\vetx(t)=[x_1(t),x_2(t),\ldots,x_n(t)]^T$ represent the state of the MV-QHNN at time $t$, that is, the component $x_i(t)$ corresponds to the state of the $i$th neuron.  Like the complex-valued Hopfield network of Jankowski \cite{jankowski96}, the state of the $i$th neuron of the MV-QHNN (of Minemoto et al.) is determined by the {\it phase-quanta} \cite{minemoto16}: Given positive integers $K_1$, $K_2$, and $K_3$, called \textit{resolution factors}, the {\it phase quanta} are defined by
\bb \label{eq:deltas} \Delta \phi = \frac{2\pi}{K_1}, \quad  \Delta \psi = \frac{\pi}{2K_2}, \quad \mbox{and} \quad \Delta \theta = \frac{\pi}{K_3}. \ee
Precisely, the state of the $i$th neuron is an unit quaternion of the form
\bb x_i(t) = e^{\phi_i(t)\ii} e^{\psi_i(t)\kk} e^{\theta_i(t)\jj},\ee
where the phase-angles belong to the following finite sets:
\begin{align*}
 \phi_i(t) \in & \left\lbrace \frac{-2\pi+(2\ell+1) \Delta \phi}{2}: \ell = 0,\ldots,K_1-1 \right\rbrace,\\
 \psi_i(t) \in & \left\lbrace \frac{-\pi/2+(2\ell+1) \Delta \psi}{2}: \ell = 0,\ldots,K_2-1 \right\rbrace, \\
 \theta_i(t) \in & \left\lbrace \frac{-\pi+(2\ell+1) \Delta \theta}{2}: \ell = 0,\ldots,K_3-1 \right\rbrace.
\end{align*}
As usual, the activation potential of the $i$th neuron at time $t$ is the quaternion defined by 
\bb \label{eq:PotActivation} v_i(t)=\sum_{j=1}^n w_{ij} x_j(t). \ee
If the activation potential has a phase-angle representation 
\bb v_i(t) = |v_i(t)| e^{\alpha_i(t) \ii} e^{\beta_i(t) \kk} e^{\gamma_i(t) \jj},\ee
then the next state $\vetx(t+\Delta t)$ of the MV-QHNN is obtained by updating the $i$th neuron according to the rule
\bb \label{eq:mupdateMine} x_i(t+\Delta t) = \begin{cases} e^{\phi_M \ii} e^{\psi_i(t) \kk} e^{\theta_i(t) \jj}, \\ \mbox{or} \\e^{\phi_i(t) \ii} e^{\psi_M \kk} e^{\theta_i(t) \jj}, \\ \mbox{or} \\e^{\phi_i(t) \ii} e^{\psi_i(t) \kk} e^{\theta_M \jj} \end{cases} \ee
where the phase-angles $\phi_M$, $\psi_M$, and $\theta_M$  are the midpoints of the arcs that contain respectively $\alpha_i(t)$, $\beta_i(t)$, and $\gamma_i(t)$. Formally, we have
\begin{align}
\label{eq:phiqsgn2M} \phi_M&=\frac{1}{2} \left( -2\pi+\Delta \phi \left( 2 \left \lfloor \frac{\pi+\alpha_i(t)}{\Delta \phi} \right \rfloor +1\right) \right),\\
\label{eq:psiqsgn2M} \psi_M&=\frac{1}{2} \left( -\frac{\pi}{2}+\Delta \psi \left( 2 \left \lfloor \frac{\frac{\pi}{4}+\beta_i(t)}{\Delta \psi} \right \rfloor +1\right) \right), \\
\label{eq:thetaqsgn2M} \theta_M&=\frac{1}{2} \left( -\pi+\Delta \theta \left( 2 \left \lfloor \frac{\frac{\pi}{2}+\gamma_i(t)}{\Delta \theta} \right \rfloor +1\right) \right).
\end{align}
Note that the angles $\phi_M$, $\psi_M$, and $\theta_M$ do not differ from $\alpha_i(t)$, $\beta_i(t)$, and $\gamma_i(t)$ by more than half of the phase quanta, i.e., the following inequalities hold true:
\bb \label{eq:H2a} |\phi_M-\alpha_i(t)|<\frac{\Delta \phi}{2}, \; |\psi_M-\beta_i(t)|< \frac{\Delta \psi}{2}, \; |\theta_M - \gamma_i(t)|<\frac{\Delta \theta}{2}.\ee
We would like to point out that, in order to avoid ambiguities, the state of the $i$th neuron remains unchanged if the activation potential $v_i(t)$ does not have an unique phase-angle representation. In other words, $x_i(t+\Delta t)=x_i(t)$ if $v_i(t) \not \in \mathcal{A}$, where $\mathcal{A}$ is the set given by \eqref{eq:SetA}. 

\begin{rmk} 
Note that \eqref{eq:mupdateMine} allows for 3 different ways to update a neuron. In order to circumvent ambiguities, we first update the phase-angle $\phi$ of all neurons of the network. Then, we update the phase-angle $\psi$ and, lastly, we update the phase-angle $\theta$, for $i=1,\ldots,n$. Furthermore, we consider $\Delta t = 1/(3n)$ if the neurons are updated asynchronously and $\Delta t =1/3$ if the neurons are updated in parallel. In both cases, the three phase-angles are updated sequentially for the whole MV-QHNN in one time unit.
\end{rmk}

The dynamic of a QHNN model is predominantly analyzed by considering the energy function
\bb \label{eq:Energy} E(\vetx) = -\frac{1}{2} \vetx^* W \vetx, \ee
where $\vetx^*$ denotes the conjugate transpose of $\vetx$ and $W$, whose entries are $w_{ij}$, is the quaternionic synaptic weight matrix. In fact, the sequence $\{\vetx(t)\}_{t \geq 0}$ produced by a QHNN is convergent if the strict inequality
\[ \Delta E(t) = E(\vetx(t+\Delta t)) -  E(\vetx(t)) < 0,\]
holds true whenever $\vetx(t+\Delta t) \neq \vetx(t)$. In this case, the time evolution of the QHNN yields a minima of \eqref{eq:Energy}. Equivalently, the network comes to rest at an equilibrium. Like the traditional Hopfield neural network, the convergence is often ensured by assuming that the neurons are updated asynchronously \cite{hopfield82,hassoun93,hassoun97}.
Also, the conditions 
\bb \label{eq:Wconditions} w_{ij} = \bar{w}_{ji} \quad \mbox{and} \quad w_{ii} \geq 0, \quad \forall i,j \in \{1,\ldots,n\}, \ee 
are usually required for the convergence of the sequence $\{\vetx(t)\}_{t \geq 0}$ produced by a QHNN model \cite{isokawa13,valle14bracis,minemoto16}.

Based on the arguments of Isokawa and collaborators \cite{isokawa08b,isokawa13}, which also hold true for the MV-QHNN given by \eqref{eq:mupdateMine}, Minemoto et al. claim that $\Delta E(t)<0$ if condition \eqref{eq:Wconditions} is satisfied \cite{minemoto16}. In the following, we address the subtle assumptions which have been wrongly accepted to show that the MV-QHNN always comes to rest at an equilibrium state. 

\subsection{Argument of Isokawa and Collaborators} \label{sec:arguments}

First, Isokawa et al. derived from \eqref{eq:Energy} the algebraic identity
\begin{equation} \label{eq:x1x2x3}
\Delta E = -(X_1-X_2)+w_{i i}(X_3-1), 
\end{equation}
where $X_1=\re{\bar{x}_i(t+\Delta t)v_i(t)}$, $X_2=\re{\bar{x}_i(t)v_i(t)}$, and $X_3=\re{\bar{x}_i(t+\Delta t)x_i(t)}$. Then, they expressed $X_1$, $X_2$, and $X_3$ using phase-angles.

The phase-angles of $x_i(t+\Delta t)$ can be obtained by shifting the phase-angles of $x_i(t)$ by multiples of the phase quanta. In mathematical terms, we have
\begin{align}
\label{eq:inta} \phi_i(t+\Delta t)&=\phi_i(t)+a \Delta \phi,\\
\label{eq:intb} \psi_i(t+\Delta t)&=\psi_i(t)+b \Delta \psi,\\
\label{eq:intc} \theta_i(t+\Delta t)&=\theta_i(t)+c \Delta \theta,
\end{align}
where $a$, $b$ and $c$ are integers. Thus, the state $x_i(t+\Delta t)$ satisfies
\[ x_i(t+\Delta t)=e^{(a\Delta \phi +\phi_i(t)) \ii}e^{(b\Delta \psi +\psi_i(t)) \kk}e^{(c\Delta \theta +\theta_i(t)) \jj}.\]
After some algebraic manipulations, Isokawa et al. obtained
\begin{align} \label{eq:X3} X_3 =& \cos(a \Delta \phi)\cos(b \Delta \psi)\cos(c \Delta \theta) \\ \nonumber & -\sin(a\Delta \phi) \sin(b\Delta \psi+2\psi_i(t))\sin(c \Delta \theta). \end{align}
Similarly, the phase-angles $\alpha_i(t)$, $\beta_i(t)$, and $\gamma_i(t)$ of the activation potential $v_i(t)$ can be obtained by shifting $\phi_i(t+\Delta t)$, $\psi_i(t+\Delta t)$, and $\theta_i(t+\Delta t)$ by $\delta \phi$, $\delta \psi$, and $\delta \theta$. In other words, 
\begin{align} 
\label{eq:shiftalpha} \alpha_i(t) &= \phi_i(t+\Delta t)+\delta \phi,\\ 
\label{eq:shiftbeta} \beta_i(t) &= \psi_i(t+\Delta t)+\delta \psi,\\ 
\label{eq:shiftgamma} \gamma_i(t) &= \theta_i(t+\Delta t)+\delta \theta,
\end{align}
and 
\[v_i(t) =  |v_i(t)| e^{\ii (\delta \phi+\phi_i(t+\Delta t))}e^{\kk (\delta \psi+\psi_i(t+\Delta t))}e^{\jj (\delta \theta+\theta_i(t+\Delta t))}.\]
Moreover, $X_1$ and $X_2$ satisfy respectively the equations
\begin{align} 
\label{eq:X1} X_1 &= |v_i(t)|(\cos(\delta \theta)\cos(\delta \psi)\cos(\delta \phi)\nonumber \\ &  -\sin(\delta \theta)\sin(2b\Delta \psi+2\psi_i(t)+\delta \psi)\sin(\delta \phi)), \end{align} 
and 
\begin{align}
\label{eq:X2}  & X_2 =|v_i(t)|(\cos(c \Delta \theta+\delta \theta)\cos(b\Delta \psi+\delta \psi)\cos(a\Delta \phi +\delta\phi)\nonumber \\ &-\sin(c\Delta \theta +\delta \theta)\sin(b\Delta \psi+2\psi_i(t)+\delta \psi)\sin(a\Delta \phi + \delta \phi)).
\end{align}

It turns out that $X_1$, $X_2$, and $X_3$ can be simplified by considering the update scheme defined by \eqref{eq:mupdateMine}. Precisely, since either $\phi_i(t+\Delta t)=\phi_i(t)$ or $\theta_i(t+\Delta t)=\theta_i(t)$, we have either $a=0$ or $c=0$ in \eqref{eq:X3}. On one hand, if $a=0$, then $X_3=\cos(c \Delta \theta)\cos(b \Delta \psi) \leq 1$. On the other hand, we have  $X_3=\cos(a \Delta \phi)\cos(b \Delta \psi) \leq 1$ if $c=0$. In both cases, the inequality $w_{ii}(X_3 - 1) \leq 0$ holds true if $w_{ii}\geq 0$.

In order to analyze \eqref{eq:X1} and \eqref{eq:X2}, the authors admit the following hypothesis:
\bb \label{eq:H1} \tag{H1} a=0 \Longleftrightarrow \delta\phi=0 \quad \mbox{and} \quad c=0 \Longleftrightarrow \delta\theta=0. \ee
As a consequence, if $a=0$ then $\delta \phi=0$ and 
\begin{align*} X_1-X_2 &=  |v_i(t)|(\cos(\delta \theta)\cos(\delta\phi) \nonumber \\&-\cos(c \Delta \theta +\delta\theta)\cos(b \Delta \psi + \delta \psi)).
\end{align*}
Similarly, if $c=0$ then $\delta \theta=0$ and 
\begin{align*} \label{eq:X1-X2c} X_1-X_2 &=  |v_i(t)|(\cos(\delta \psi)\cos(\delta\phi)\nonumber \\ &-\cos(b \Delta \psi +\delta\psi)\cos(a \Delta \phi + \delta \phi)).
\end{align*}
In both cases, the difference $X_1-X_2$ is positive under the additional hypothesis inspired by \eqref{eq:H2a}:
\bb \label{eq:H2} \tag{H2} |\delta \phi|<\Delta \phi, \quad |\delta \psi|<\Delta \psi, \quad \mbox{and} \quad |\delta \theta|<\Delta \theta. \ee
Concluding, if both \eqref{eq:H1} and \eqref{eq:H2} hold true, then the inequality $\Delta E<0$ also holds and the MV-QHNN comes to rest at an equilibrium.

\subsection{A Detailed Counterexample for the MV-QHNN}

It turns out that hypothesis (H1) is false because the activation potential $v_i(t)$ does not depend on the update scheme of the MV-QHNN model. Also, although the inequalities \eqref{eq:H2a} hold true, the phase shifts $\delta \phi$ or $\delta \theta$ may take arbitrary values if either $\phi_i(t+\Delta t)=\phi_i(t)$ or $\theta_i(t+\Delta t)=\theta_i(t)$, as prescribed by \eqref{eq:mupdateMine}. As a consequence, we cannot assert that $\Delta E<0$ if the $i$th neuron changes its state. Also, we are not able to ensure that the sequence produced by the network is convergent. 

\begin{exmp} \label{ex:Minemoto1}
Consider $K_1=K_2=K_3=2$ and the synaptic weight matrix 
\bb \label{eq:W2}  W = \begin{bmatrix} 0 & 5+\ii+7\jj+2\kk \\ 5-\ii-7\jj-2\kk & 0 \end{bmatrix} \in \mathbb{H}^{2 \times 2}. \ee
Note that $W$ satisfies the usual quaternionic conditions \eqref{eq:Wconditions}. Also, consider the initial state 
\bb \label{eq:x0} \vetx(0)=\begin{bmatrix} e^{-\frac{\pi}{2}\ii}e^{-\frac{\pi}{8}\kk}e^{-\frac{\pi}{4}\jj} \\ e^{-\frac{\pi}{2}\ii}e^{-\frac{\pi}{8}\kk}e^{-\frac{\pi}{4}\jj}\end{bmatrix}.\ee 
Evidently, the phase-angles of $x_1(0)$ are 
\[ \phi_1(0)=-\frac{\pi}{2}, \quad \psi_1(0)=-\frac{\pi}{8}, \quad \mbox{and} \quad \theta_1(0)=-\frac{\pi}{4}.\]
The energy of MV-QHNN at $\vetx(0)$ is
\[ E(\vetx(0)) = -\frac{1}{2} \vetx^*(0) W \vetx(0) = -5.\] 
The activation potential of the first neuron at time $t=0$ is \[v_1(0)=-0.1121+1.577\ii-5.207\jj+7.028\kk,\] and its phase-angles are 
\bb \label{eq:anglesv} \alpha_1(0)=2.1939, \; \beta_1(0)=0.09455, \; \mbox{and} \; \gamma_1(0)=1.4181.\ee
From \eqref{eq:phiqsgn2M}-\eqref{eq:thetaqsgn2M}, we obtain 
\bb \label{eq:angles_M} \phi_M=\frac{\pi}{2}, \quad \psi_M=\frac{\pi}{8}, \quad \mbox{and} \quad \theta_M=\frac{\pi}{4}.\ee
According to \eqref{eq:mupdateMine}, if we update the phase-angle $\phi$ of the first neuron, we obtain
$x_1(1/6) = e^{\frac{\pi}{2}\ii} e^{-\frac{\pi}{8}\kk} e^{-\frac{\pi}{4}\jj}$.
Note that the phase-angles 
\[\phi_1(1/6)=\frac{\pi}{2}, \quad \psi_1(1/6)=-\frac{\pi}{8}, \quad \mbox{and} \quad \theta_1(1/6)=-\frac{\pi}{4},\] of $x_1(1/6)$
satisfy \eqref{eq:inta}-\eqref{eq:intc} with the integers:
\[ a=1, \quad b=0, \quad \mbox{and} \quad c=0.\]
Also, the shifts $\delta \phi$, $\delta \psi$, and $\delta \theta$ in \eqref{eq:shiftalpha}-\eqref{eq:shiftgamma} are given by
\[ \delta \phi=0.6231,\quad \delta \psi=0.4873, \quad \mbox{and} \quad \delta\theta = 2.2035. \]
Note that $c=0$ because $\theta_1(1/6)=\theta_1(0)$, but $\delta \theta \neq 0$, which contradicts (H1). In addition, $\delta \theta=2.2035 > \frac{\pi}{2}=\Delta \theta$, violating (H2).
Finally, the energy of the network at the state vector
$\vetx(1/6) = [e^{\frac{\pi}{2}\ii}e^{-\frac{\pi}{8}\kk}e^{-\frac{\pi}{4}\jj},e^{-\frac{\pi}{2}\ii}e^{-\frac{\pi}{8}\kk}e^{-\frac{\pi}{4}\jj}]^T$ 
is
\[ E(\vetx(1/6)) = -\frac{1}{2} \vetx^*(1/6) W \vetx(1/6) = 5.\]
Thus, the variation of the energy of the network from $t=0$ to $t=1/6$ is $\Delta E = E(\vetx(1/6))-E(\vetx(0)) = 10>0$.
By updating the phase-angle $\phi$ of the second neuron, we obtain 
$\vetx(1/3) = [e^{\frac{\pi}{2}\ii}e^{-\frac{\pi}{8}\kk}e^{-\frac{\pi}{4}\jj},e^{\frac{\pi}{2}\ii}e^{-\frac{\pi}{8}\kk}e^{-\frac{\pi}{4}\jj}]^T$ and, the energy of the MV-QHNN at this state is $E(\vetx(1/3)) = -5$.
Proceeding in a similar manner, the asynchronous MV-QHNN does not reach a stationary state. In fact, we obtain a periodic sequence of quaternionic state vectors. Similarly, the parallel MV-QHNN also yields a periodic sequence of quaternionic vectors. The energy produced by the MV-QHNN of Minemoto using either asynchronous or parallel updates is shown in Fig. \ref{fig:Example1}. 

\begin{figure}
\[ \includegraphics[width=1\columnwidth]{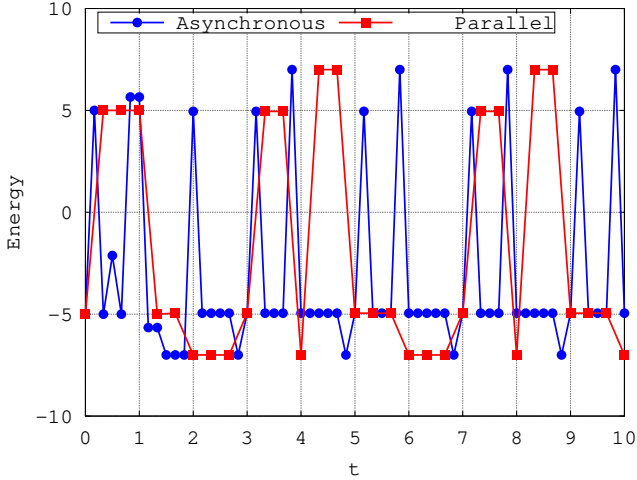}\]
\caption{Energy of the asynchronous and parallel MV-QHNN of Minemoto.} \label{fig:Example1}
\end{figure}
\end{exmp}

\subsection{Modified MV-QHNN Model}

In Example \ref{ex:Minemoto1}, we pointed out that the update scheme \eqref{eq:mupdateMine} violates the hypothesis \eqref{eq:H2}. It turns out, however, that \eqref{eq:H2} holds true if we consider the following update scheme in which the three phase-angles are updated simultaneously:
\bb \label{eq:mupdateMine2} x_i(t+\Delta t) = e^{\phi_M \ii} e^{\psi_M \kk} e^{\theta_M \jj}, \ee
where $\phi_M$, $\psi_M$, and $\theta_M$ are defined by \eqref{eq:phiqsgn2M}-\eqref{eq:thetaqsgn2M}, respectively. On the downside, we cannot simplify the terms $X_1$, $X_2$, and $X_3$ in \eqref{eq:x1x2x3}. In fact, contrary to what we conjectured in \cite{valle16wcci}, we cannot guarantee $\Delta E<0$ yet. The following example confirms this remark. 

\begin{rmk}
Since the three phase-angles of a neuron are updated simultaneously, we refer to the model defined by \eqref{eq:mupdateMine2} as MV-QHNN3. Again, we assume that all the neurons of the MV-QHNN3 are updated in one time unit. Thus, $\Delta t = 1/n$ if the neurons are updated asynchronously and $\Delta t =1$ if the neurons are updated in parallel. 
\end{rmk}

\begin{exmp} \label{ex:Minemoto2}
Consider $K_1=K_2=K_3=2$ and let $W$ and $\vetx(0)$ be the quaterionic synaptic weight matrix and initial state given respectively by \eqref{eq:W2} and \eqref{eq:x0}. By considering \eqref{eq:mupdateMine2} instead of \eqref{eq:mupdateMine}, the first neuron becomes 
\[x_1(1/2)=e^{\phi_M \ii} e^{\psi_M \kk} e^{\theta_M \jj} = e^{\frac{\pi}{2}\ii}e^{\frac{\pi}{8}\kk}e^{\frac{\pi}{4}\jj}, \]
where $\phi_M$, $\psi_M$, and $\theta_M$ are given by \eqref{eq:angles_M}. The energy of the network in the quaternionic state vector $\vetx(1/2) = [e^{\frac{\pi}{2}\ii}e^{\frac{\pi}{8}\kk}e^{\frac{\pi}{4}\jj}, e^{-\frac{\pi}{2}\ii}e^{-\frac{\pi}{8}\kk}e^{-\frac{\pi}{4}\jj}]^T$ is
\[ E(\vetx(1/2)) = -\frac{1}{2} \vetx^*(1/2) W \vetx(1/2) = -7.\]
Although the energy decreased from $t=0$ to $t=1/2$, the asynchronous MV-QHNN3 model does not settle down into a stationary state. In fact, this network yields a periodic sequence. Fig.  \ref{fig:Example2} shows the energy of the asynchronous MV-QHNN3 described by \eqref{eq:mupdateMine2}. This figure also shows the energy of the MV-QHNN3 obtained by updating the neurons in parallel, which also produces a periodic sequence.

\begin{figure}
\[ \includegraphics[width=1\columnwidth]{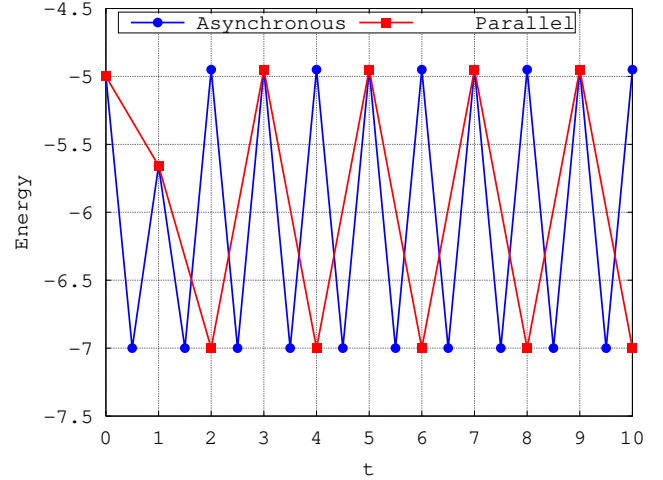}\]
\caption{Energy of the asynchronous and parallel MV-QHNN3 model. } \label{fig:Example2}
\end{figure}
\end{exmp} 

\subsection{Large Resolution Factors} \label{subsec:LargeResolution}

Although we provided an example in which the energy of the MV-QHNN3 described by \eqref{eq:mupdateMine2} increases, this network will probably come to rest at an equilibrium state if the resolution factors $K_1$, $K_2$, and $K_3$ are sufficiently large. Indeed, consider the following: 
\begin{itemize}
 \item $A_1$ is the angle between $x_i(t+\Delta t)$ and $v_i(t)$,
 \item $A_2$ the angle between $x_i(t)$ and $v_i(t)$, 
 \item $A_3$ the angle between $x_i(t+\Delta t)$ and $x_i(t)$.
\end{itemize}
From \eqref{eq:cossenos}, the terms $X_1$, $X_2$, and $X_3$ in the algebraic identity $\Delta E = -(X_1-X_2)+w_{i i}(X_3-1)$ satisfy
\begin{align}
\label{eq:x1cauchy} X_1 &=\re{\bar{x}_i(t+\Delta t)v_i(t)}=|v_i(t)|\cos(A_1), \\ 
\label{eq:x2cauchy} X_2 &=\re{\bar{x}_i(t)v_i(t)}=|v_i(t)|\cos(A_2),\\
\label{eq:x3cauchy} X_3 &=\re{\bar{x}_i(t+\Delta t)x_i(t)}=\cos(A_3).
\end{align}
Note that, if the $i$th neuron of the network changes its state from time $t$ to $t + \Delta t$, then $x_i(t+\Delta t) \neq x_i(t)$. Thus, $X_3=\cos(A_3)<1$ because $x_i(t+\Delta t)$ and $x_i(t)$ are not parallel. Furthermore, the difference $X_1-X_2$ satisfies
\bb X_1-X_2 = |v_i(t)|\big( \cos(A_1)-\cos(A_2)\big).\ee
Now, if $K_1,K_2,K_3 \rightarrow \infty$, then $\Delta\phi, \Delta\psi,\Delta\theta \rightarrow 0$. Moreover, $\phi_M \rightarrow \alpha_i(t)$, $\psi_M \rightarrow \beta_i(t)$, and $\theta_M \rightarrow \gamma_i(t)$. As a consequence, we have $A_1 \rightarrow 0$. In other words, if the resolution factors $K_1$, $K_2$, and $K_3$ are sufficiently large, then $x_i(t+\Delta t)$ and $v_i(t)$ are almost parallel vectors and, consequently, $\cos(A_1) \rightarrow 1$. In contrast, $A_2$ usually does not tend to $0$ and, thus, the inequality $\cos(A_2)<1$ often holds. 
Concluding, if the resolution factors are sufficiently large, the variation of energy of the network should satisfy
\[ \Delta E \rightarrow |v_i(t)|(\cos(A_2)-1)+w_{ii}(\cos(A_3)-1)<0.\]
These arguments leads to the {\it continuous-valued Quaternionic Hopfield Neural Network} (CV-QHNN), which corresponds to the limit $K_1,K_2,K_3 \to \infty$. 

\section{Continuous-Valued Quaternionic Hopfield Neural Network} \label{sec:CV-QHNN}

The CV-QHNN, proposed by Valle \cite{valle14bracis} and further investigated by Kobayashi\footnote{We would like to point out that the CV-QHNN corresponds to the left-QHNN model in \cite{Kobayashi16a}.} \cite{Kobayashi16a}, corresponds to the limit of the modified MV-QHNN described by \eqref{eq:mupdateMine2} when the resolution factors tend to infinity. The evolution of the CV-QHNN is defined by 
\bb \label{update:CV-QHNN} x_i(t+\Delta t) = \begin{cases} v_i(t)/|v_i(t)|, & v_i(t) \neq 0, \\ x_i(t), &  \mbox{otherwise}. \end{cases} \ee
In few words, the next state of the $i$th neuron is obtained by normalizing its activation potential to length one. Thus, it can be easily implemented and analyzed. In particular, it does not require the phase-angle representation of quaternions. Notwithstanding, if we know the phase-angle representation $v_i(t)=|v_i(t)| e^{\alpha_i(t) \ii} e^{\beta_i(t) \kk} e^{\gamma_i(t) \jj}$, the next state of the $i$th neuron can be alternatively determined by \bb \label{eq:CV-QHNN2} x_i(t) = e^{\alpha_i(t) \ii} e^{\beta_i(t) \kk} e^{\gamma_i(t) \jj}.\ee
We would like to point out that $\sigma$ can be viewed as the quaternionic version of continuous-valued complex activation function proposed by Aizenberg et al. \cite{aizenberg05}. This kind of activation function has also been considered by Kobayashi in his hyperbolic Hopfield neural network model \cite{kobayashi13}. 

The sequence produced by \eqref{update:CV-QHNN}, in an asynchronous update mode, is convergent for any initial state $\vetx(0) \in \mathbb{S}^n$ if the synaptic weights satisfy $w_{ij}=\bar{w}_{ji}$ and $w_{ii} \geq 0$ for any $i,j \in \{1,\ldots,n\}$ \cite{valle14bracis}. Precisely, the energy of the asynchronous CV-QHNN always decreases if a neuron changes its states. In general terms, we derive this result from \eqref{eq:x1x2x3} by noting that \eqref{eq:x1cauchy}-\eqref{eq:x3cauchy} hold true with $\cos(A_1)=1$, $\cos(A_2)<1$, and $\cos(A_3)<1$. Although we can ensure the convergence of the sequence produced by the asynchronous CV-QHNN, nothing can be said about the CV-QHNN using parallel dynamic. The following example, which is similar to Examples \ref{ex:Minemoto1} and \ref{ex:Minemoto2}, confirms these remarks.

\begin{rmk}
Accordingly, we assume that all the neurons of the CV-QHNN are updated in one time unit. Thus, $\Delta t = 1/n$ if the neurons are updated asynchronously and $\Delta t =1$ if the neurons are updated in parallel.
\end{rmk}

\begin{exmp} \label{ex:Valle1}
Consider $K_1=K_2=K_3=2$. Let the synaptic weight matrix $W$ and the initial state $\vetx(0)$ be given respectively by \eqref{eq:W2} and \eqref{eq:x0}. Using asynchronous update, we obtain from \eqref{update:CV-QHNN} the quaternionic vector 
\[ \vetx(1/2) = \begin{bmatrix} -0.01261 + 0.1774\ii - 0.5858\jj + 0.7907\kk \\ -0.2706 - 0.6533\ii - 0.2706\jj + 0.6533\kk \end{bmatrix},\]
which is a stationary state of the network. Furthermore, it is not hard to verify that 
\[ \Delta E = E(\vetx(1/2))-E(\vetx(0)) = -8.888+5=-3.888.\]
In contrast, if we use parallel update, \eqref{update:CV-QHNN} yields the vector
\[ \vetx(1) = \begin{bmatrix} -0.01261 + 0.1774\ii - 0.5858\jj + 0.7907\kk \\ -0.2918 - 0.9124\ii + 0.2814\jj - 0.05567\kk \end{bmatrix},\]
whose energy value is the same as that of the initial state, i.e., $E(\vetx(1)) = -5$. Proceeding in a similar manner, we obtain
\[ \vetx(2) = \begin{bmatrix} -0.2706 - 0.6533\ii - 0.2706\jj + 0.6533\kk \\ -0.2706 - 0.6533\ii - 0.2706\jj + 0.6533\kk \end{bmatrix},\]
which, despite the algebraic representation, is equal to the initial state $\vetx(0)$. Therefore, the parallel CV-QHNN oscillates between the quaternionic state vectors $\vetx(0)$ and $\vetx(1)$ without changing the energy value. Fig. \ref{fig:Example3} shows the energy of the CV-QHNN using both asynchronous and parallel update modes. 

\begin{figure}
\[ \includegraphics[width=1\columnwidth]{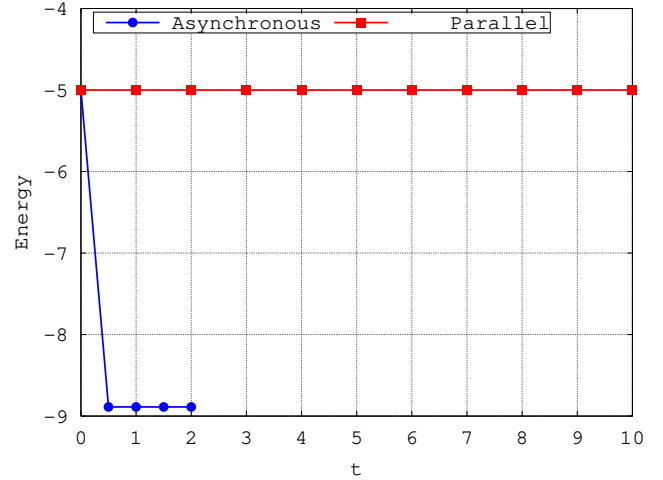}\]
\caption{Energy of the asynchronous and parallel CV-QHNN model. } \label{fig:Example3}
\end{figure}
\end{exmp} 

As pointed out previously, the CV-QHNN model can be obtained from the MV-QHNN3 by taking the limit $K_1,K_2,K_3 \to \infty$. Since the asynchronous CV-QHNN always settle down into an equilibrium, the asynchronous MV-QHNN3 must also come to rest at equilibrium state if the resolution factors are sufficiently large. The following example addresses this issue. Moreover, in contrast to the previous examples, we consider quaternionic networks with 100 neurons. 

\begin{exmp}
In order to evaluate the convergence of sequence produced by the QHNN models, we proceeded as follows:
We first generated a $100 \times 100$ quaternionic matrix $R$ with entries $r_{ij}=\quat{\mathtt{randn}}$, where $\mathtt{randn}$ yields a random scalar drawn from the standard normal distribution. Then, we computed $U=\frac{1}{2}(R+R^*)$, where $R^*$ denotes the conjugate transpose of $R$, and defined the quaternionic synaptic weight matrix by 
\[ W = U - \mathtt{diag}(u_{11},u_{22},\ldots,u_{100,100}),\]
where $\mathtt{diag}(u_{11},\ldots,u_{100,100})$ is the diagonal matrix composed of the diagonal elements of $U$. Note that $W$ satisfies $w_{ij}=\bar{w}_{ji}$ and $w_{ii}=0$ for all any indexes $i$ and $j$. Moreover, the real and imaginary parts of $w_{ij}$ are normally distributed random numbers. Given the resolution factors $K_1$, $K_2$, and $K_3$, we also defined the initial quaternionic vector $\vetx(0)$ by means of the equation
$x_i(0)=e^{\phi_i(0)\ii}e^{\psi_i(0)\kk} e^{\theta_i(0)\jj}$, for all $i=1,\ldots,n$, where 
\begin{align*}
\phi_i(0) &=\frac{1}{2}\left(-2\pi+\Delta \phi (2\mathtt{randi}(K_1)+1)\right), \\ 
\psi_i(0) &= \frac{1}{2}\left(-\frac{\pi}{2}+\Delta \psi (2\mathtt{randi}(K_2)+1)\right), \\ 
\theta_i(0) &= \frac{1}{2}\left(-\pi+\Delta \theta (2\mathtt{randi}(K_3)+1)\right),
\end{align*}
and $\mathtt{randi}(K_\ell)$ yields an integer between $0$ and $K_\ell-1$. Fig. \ref{fig:probability} shows the probability of a randomly generated QHNN model to settle down into an equilibrium state in at most $1000$ iterations, that is, we allowed the network to evolve while $t \leq 1000$. The probabilities have been computed by repeating the procedure 100 times for each resolution factor. We would like to point out that the synaptic weight matrix $W$ and the initial state $\vetx(0)$ were the same for all QHNN models.
\begin{figure}
\[ \includegraphics[width=1\columnwidth]{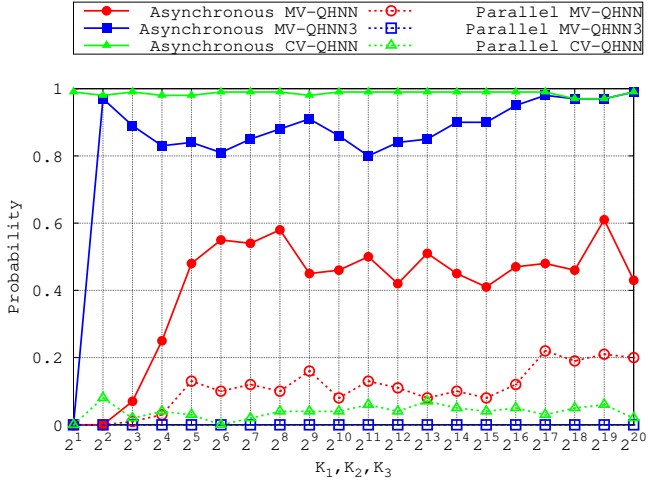}\]
\caption{Probability of the QHNN models to settle down into an equilibrium by the resolution factors.} \label{fig:probability}
\end{figure}
Note that the probability of a parallel model to reach a stationary state is always less than or equal to the probability of the corresponding asynchronous model. In particular, the parallel MV-QHNN3 failed to settle down into an equilibrium in all simulations for any resolution factors. In contrast, the probability of the asynchronous MV-QHNN3 to settle down into a stationary state is greater than $0.80$ for all $K_1,K_2,K_3 \geq 4$. Furthermore, in agreement with Section \ref{subsec:LargeResolution}, the asynchronous MV-QHNN3 coincides with the asynchronous CV-QHNN model if $K_1$, $K_2$, and $K_3$ are sufficiently large, namely, $K_1,K_2,K_3 \geq 2^{17}$. The asynchronous MV-QHNN of Minemoto often failed to come to rest at an equilibrium state even for large resolution factors. Finally, although the asynchronous CV-QHNN model eventually failed to reach a stationary state, we confirmed that this network tended monotonically to equilibrium.
\end{exmp}

\section{Concluding Remarks} \label{sec:concluding}
Multivalued quaternionic Hopfield neural networks (MV-QHNNs) represent an elegant generalization of the Hopfield network using unit quaternions. In this paper, we first addressed the convergence of the sequence produced by the MV-QHNN of Minemoto \cite{minemoto16}. Then, we considered a modification of the MV-QHNN model, referred to as MV-QHNN3, in which all phase-angle are updated simultaneously. Contrary to what was believed, these networks do not always comes to rest at a stationary state. Indeed, we provided a simple example in which the MV-QHNN of Minemoto as well as the MV-QHNN3 oscillates forever using both parallel and asynchronous update mode. Furthermore, we pointed out exactly which were the wrong subtle assumptions used to show the convergence of sequence defined by the MV-QHNN model \cite{isokawa08b,isokawa13}. We believe this is an important theoretical issue because MV-QHNNs have been used as the basis for many other quaternionic associative memory models \cite{minemoto15a}. Moreover, we expect to instigate further research on the dynamics of multivalued quaternionic recurrent neural networks.  

Besides the multivalued models, in this paper we also addressed the continuous-valued Hopfield neural network (CV-QHNN) \cite{valle14bracis,Kobayashi16a}. The CV-QHNN can be implemented and analyzed more easily than the MV-QHNN models. Although the CV-QHNN corresponds to a limit case of a MV-QHNN, the asynchronous CV-QHNN always produces a convergent sequence under the usual conditions on the synaptic weights. Like the traditional bipolar Hopfield network, the parallel CV-QHNN may oscillate forever.




\end{document}